\documentclass{article}

\PassOptionsToPackage{numbers, compress}{natbib}
\usepackage[final]{nips_2017}

\usepackage[utf8]{inputenc} % allow utf-8 input
\usepackage[T1]{fontenc}    % use 8-bit T1 fonts
\usepackage{hyperref}       % hyperlinks
\usepackage{url}            % simple URL typesetting
\usepackage{booktabs}       % professional-quality tables
\usepackage{amsfonts}       % blackboard math symbols
\usepackage{amssymb}
\usepackage{amsmath}
\usepackage{nicefrac}       % compact symbols for 1/2, etc.
\usepackage{microtype}      % microtypography
\usepackage{graphicx}
\usepackage{subfig}

\title{Automated Knee X-ray Report Generation}

\author{
  Aydan Gasimova\\
  Biomedical Image Analysis Group\\
  Imperial College London\\
  \And
  Giovanni Montana \\
  Imaging and Biomedical Engineering \\
  King's College London \\
  \And
  Daniel Rueckert \\
  Biomedical Image Analysis Group\\
  Imperial College London\\
}

\begin{document}
% \nipsfinalcopy is no longer used

\maketitle

\begin{abstract}
Gathering manually annotated images for the purpose of training a predictive model is far more challenging in the medical domain than for natural images as it requires the expertise of qualified radiologists. We therefore propose to take advantage of past radiological exams (specifically, knee X-ray examinations) and formulate a framework capable of learning the correspondence between the images and reports, and hence be capable of generating diagnostic reports for a given X-ray examination consisting of an arbitrary number of image views. We demonstrate how aggregating the image features of individual exams and using them as conditional inputs when training a language generation model results in auto-generated exam reports that correlate well with radiologist-generated reports.
\end{abstract}

\section{Introduction}
The burden on diagnostic radiologists has been increasing quite rapidly with the advancement of a variety of modern imaging modalities, which, while allowing for higher resolution images in 3D and even 4D, dramatically increase the complexity of the diagnostic process. It has become common for radiologists to rely on various image analysis and automated decision support systems to facilitate the interpretation process. These computer aided diagnostic systems, or CAD, have been able to make great advances with the help of machine learning algorithms and a large amount of clinical imaging data, and have been successfully tested in clinical settings \cite{freer2001, ramirez2013, Litjens2015, kobayashi2016}.
% available through a hospital's picture archiving and communication system (PACS), 

Many of these machine learning algorithms are supervised learning approaches that require large amounts of annotated image data for training. Gathering suitable data is especially challenging in the medical domain as it is incredibly time consuming for radiologists to generate ground-truths of the standard and volume required for training a predictive model. An alternative approach is to use past clinical images and corresponding radiological reports available through a hospital's picture archiving and communication system (PACS); the advantage being that, although this data is largely unstructured (free text), it is available to us in high volumes and removes the need for manual annotation.

Hence, it is necessary to develop an appropriate learning framework: one that can take advantage of past radiological examinations and their corresponding textual reports, and formulate a method for prediction that can best expedite the diagnostic process. An additional motivation to learning from clinical reports is that reports provide a much richer context to the pathology than a single class label, such as severity and location.
%, which provides the radiologist with a level of interpretability to the predictive model.

Here we present a model for captioning knee X-ray exams that is capable of generating radiological reports summarising present pathologies and the contexts within which they are presented. A key contribution is that the proposed model is able to handle an arbitrary number of input image views (which is typical in the context of knee X-ray exams) and is able to learn from free-text reports. The model is based on the neural image caption (NIC) model of \cite{Vinyals2015}, where a word-level language generation model is conditioned on image features. In our implementation, the X-ray image features for each input image are derived from a pre-trained classification network, and aggregated into a fixed-sized input to the language model.

\section{Related Work}
In the field of computer vision, the use of human generated visually descriptive text to infer the contents of an image has primarily been applied to image caption generation. Earlier models relied on linking template-based language models to objects and spatial contexts in the image \cite{Farhadi2010, Kulkarni2013, Fidler2013}. More recently, interest has moved to the combined potential of convolutional neural networks (CNNs) and recurrent neural networks (RNNs) for describing images using natural language \cite{Kiros2014b, Vinyals2015, Karpathy2015, Chen2015, Xu2015}. The advantage of using neural networks for caption generation is that the model is not constricted by hard-coded language templates and is able to learn more freely from the training data.

By contrast, using clinical reports to perform automated pathology detection in medical images has only been explored in a few studies \cite{Shin2016a, Shin2016b, Zhang2017MDnet, Zhang2017TandemNet}. In addition, these models are limited to learning from carefully curated reports, either medical subject heading annotations (MeSH\textsuperscript{\textregistered} \cite{lipscomb2000}) \cite{Shin2016a, Shin2016b}, or templated reports made specifically for the purpose of training \cite{Zhang2017MDnet, Zhang2017TandemNet}, which are both very time consuming for radiologists/pathologists to create. 

To our knowledge, this is the first study exploring the use of past medical examinations and their corresponding raw reports in order to develop a predictive model for an arbitrary number of input radiological images.

\section{Dataset}
The knee X-ray dataset has been extracted from the PAC system of St Thomas Hospital (part of Guys and St Thomas NHS Foundation Trust) and has been fully anonymised to remove sensitive patient information. It comprises a total of 330 knee X-ray exams collected over the years 2015 and 2016. Each exam consists of a textual report and one or more X-ray images (left/right knees or both, taken from different views: anteroposterior (AP), lateral (L) and skyline (S), and different positions: weight-bearing (WB) and non-weight-bearing(nonWB)). The most common exam consists of both AP and L views of left and right knees separately, making up 42\% of total exams. The reports vary in length between 2 and 145 words, with an average of 30 and standard deviation\ of 18.7; and between 1 and 16 sentences with an average of 2.7 per report. The X-ray images vary in sizes between $420 \times 650 \times 3$ and $3056 \times 3056 \times 3$.
% A typical exam is presented in Figure \ref{fig:sample_exam}. 

Prior to using the raw reports for training an image captioning model, they were cleaned to remove phrases not directly describing the images, for instance, instructions to compare to past exams (which are not available to us due to the anonymisation process). These phrases do not allow any inferences about current images, so are not useful for finding image-text correspondence. In addition, common `stopwords' (such as `and', `the') were removed, and all punctuation was replaced with either commas or full stops where appropriate. All report cleaning was done using regular expressions. Exams were randomly split into 80\%-10\%-10\% for training, testing and validation respectively.

\section{Medical Image Report Generation Model}
\begin{figure}[h]
  \centering{
    \includegraphics[width=\textwidth]{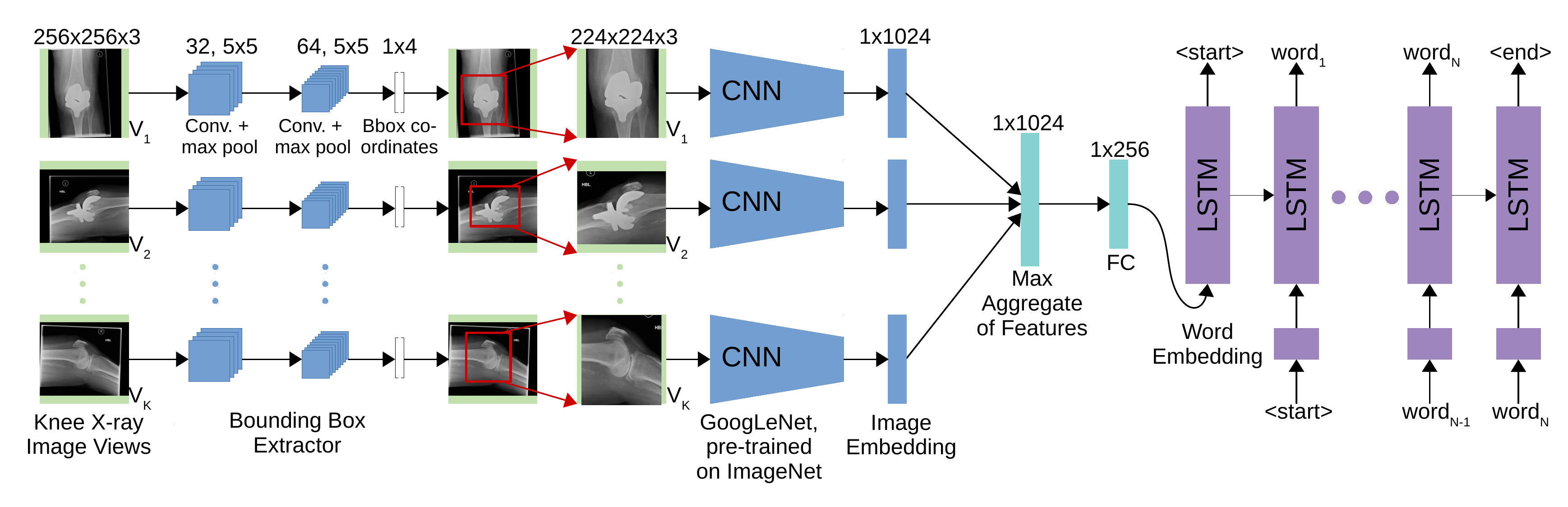}
  }
  \caption{Medical image report generation network.}
  \label{fig:caption_model}
\end{figure}

\subsection{Image Modelling}
We adopt the GoogLeNet \cite{szegedy2015} CNN architecture, pre-trained on the ImageNet dataset \cite{deng2009}, and extract image features of each X-ray image view ($V_1$ - $V_K$) from the last spatial average pooling layer ($\mathbb{R}^{1024}$). The maximum value of each feature is aggregated across the exam images to create a fixed-size input to the RNN of dimension $\mathbb{R}^{1024}$, which is then passed through a fully connected layer in order to reduce the dimension to $\mathbb{R}^{256}$, equal to the RNN input size.

\subsection{Language Modelling}
We use a recurrent neural network, specifically the \textbf{Long Short-Term Memory} (LSTM) implementation proposed in \cite{hochreiter1997}, to model the report sequence. LSTM models are widely used in machine translation \cite{sutskever2014, cho2014, luong2015} and natural image and video captioning \cite{Kiros2014b, donahue2015, Xu2015} due to their ability to capture long-term dependencies, and to reduce the problem of vanishing gradients in vanilla RNNs. Each LSTM unit has three sigmoid gates to control the internal state: `input', `output' and `forget'. At each time step, the gates control how much of the previous time steps is propagated through to determine the output. For an input word sequence $\{x_1, \dots, x_n \}$, the internal hidden state $h_t \in \mathbb{R}^{256}$ and memory state $m_t \in \mathbb{R}^{256}$ are updated as follows:
$$ h_t = f_t \odot h_{t-1} + i_t \odot \tanh(W^{(hx)}x_t + W^{(hm)}m_{t-1})$$
$$ m_t = o_t \odot \tanh(h_t)$$
where $ x_t \in \mathbb{R}^D$ is the word embedding, $W^{(hx)}$ and $W^{(hm)}$ are the trainable weight parameters, and $i_t$, $o_t$ and $f_t$ are the input, output and forget gates respectively.

In order for the sequence generation model to be conditioned on the set of input images, the combined images features for each exam are input at time step $t=0$, and the words in the report input at consequent time steps. The complete network, including a bounding box (BBox) regressor to extract the knee joint, is illustrated in Figure \ref{fig:caption_model}.

\section{Training}
For training the report generation model, the LSTM is unrolled up to 33 time steps (1 for image features, 1 per start and end token, and 30 for the average number of words in the reports). The images were rescaled and padded (preserving the aspect ratio) to $224\times 224$. Words were one-hot encoded, and ones that occurred at a frequency \textless 5 were removed.

To improve the X-ray image features, a simple CNN BBox regressor was built and trained on a subset of the training images and ground-truth BBoxes manually created for this purpose (231 image-BBox pairs). In addition, the training set was augmented eight-fold by random cropping the original images from $256\times 256$ to $224\times 224$, (in the case of the BBox images, small, random translations (between -5 and 5 in \textit{x} and \textit{y}) were applied to the BBox of each image), flipping the images along the vertical axis, and shuffling the sentences in the reports.

We then trained the LSTM model for report generation conditioned on the combined image features by minimising the negative log-likelihood between the output and true sequence:
$$ L(S,I) = -\sum_{t=0}^{N} \log p(P_t = T_t | \text{CNN($I$)}, P_0 \ldots P_{t-1}) $$

where $p$ is the probability that the predicted word $P_t$ equals the true word $T_t$ at time step $t$ given image features CNN($I$) and previous words $P_0 \ldots P_{t-1}$, and $N$ is the LSTM sequence length. At training time, loss is minimised over the training set using stochastic gradient descent (batch size 20, learning rate $1\times 10^{-5}$), and parameters are updated using Adam \cite{kingma2014} optimisation.

\section{Evaluation}
Reports are generated on the test set by max-aggregating the image features of an exam, using it as the first word input into the LSTM, and sampling consequent words. The quality of the generated reports was evaluated by measuring BLUE \cite{papineni2002} and METEOR \cite{banerjee2005} scores averaged over all the reports, which are a modified form of n-gram precision commonly used for evaluating image captioning and machine translation as they maintain high correlation with human judgement. As a comparison, we trained the model on single image inputs, duplicating the reports for each image in an exam. We also evaluated whether the performance improved if we narrow the input to the CNN to a BBox surrounding the knee joint.

\begin{figure}[h]
    \centering
    \begin{tabular}{@{}c@{}}
        \centering
        \includegraphics[width=0.5\textwidth]{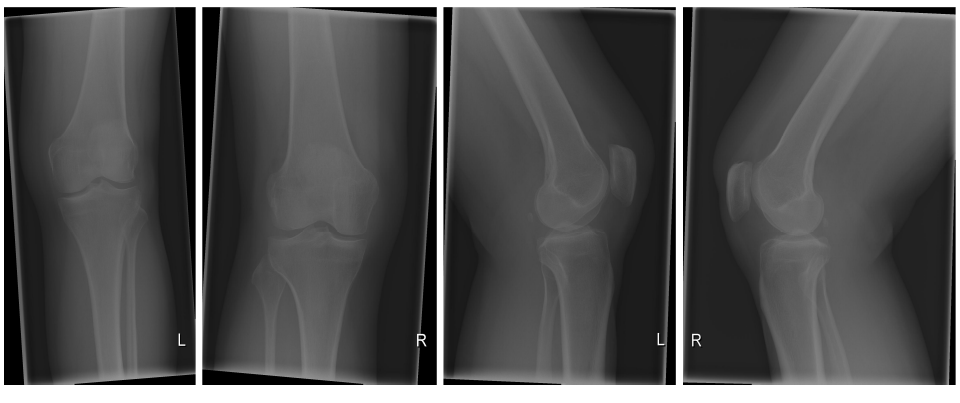}
    \end{tabular}
    ~
    \begin{tabular}{@{}c@{}}
        \centering
        \noindent\fbox{\scriptsize \parbox{0.4\textwidth}{\textbf{Original:} Joint spaces articular surfaces appear preserved. Significant degenerative erosive change seen.
        \\
        \\
        \textbf{Good prediction (B1=87.5):} Joint spaces articular surfaces appear preserved bilaterally.
        \\
        \\
        \textbf{Poor prediction (B1=28.6):} Joint space narrowing medial compartments bilaterally. }}
    \end{tabular}
    
    \caption{Sample knee X-ray exam with corresponding original report (with common stopwords removed) and two sampled predictions generated by the model trained on max-aggregates of image features.}
    \label{fig:sample_exam}
\end{figure}

The report generation model performed better when trained on a combination of image features than on single images as the conditional input (see Table \ref{tab:results}), suggesting that all images are required, in part, in order to make the pathology assessment. This is in line with expectations since pathologies may only be present in one leg, or may only be seen from a particular view. Contrary to expectation, bounding box extraction did not improve the performance of report generation as it may have instead removed pertinent image features. A sample exam with the original report and two auto-generated reports of contrasting BLEU-1 scores is presented in Figure \ref{fig:sample_exam}.

\begin{table}[]
\centering
\begin{tabular}{@{}lllllllllll@{}}
\toprule
& \multicolumn{2}{l}{BLEU-1} & \multicolumn{2}{l}{BLEU-2} & \multicolumn{2}{l}{BLEU-3} & \multicolumn{2}{l}{BLEU-4} & \multicolumn{2}{l}{METEOR} \\ \midrule
& tr           & te          & tr           & te          & tr           & te          & tr           & te          & tr           & te\\
Baseline, single image input & 42.2         & 33.2        & 13.3         & 5.7         & 3.8          & 1.9         & 1.3          & 1.1 
& 26.7 & 22.2       \\
Max-aggr. of image features  & 60.7         & 40.4        & 32.6         & 10.1        & 19.4         & 2.6         & 12.3         & 1.2            & 41.1     & 35.7\\
Max-aggr.+BBoxes            & 38.9         & 37.4        & 11.3         & 7.1        & 3.4          & 1.1         & 1.2          & 0.2 
& 28.3 & 28.9 \\ \bottomrule
\end{tabular}
\bigskip
\caption{\label{tab:results}BLEU n-gram and METEOR scores evaluated on report generation model trained on single image inputs (Baseline), max-aggregate of image features, and max-aggregate of image features extracted from BBox detections. Evaluated on training (tr) and test (te) data.}
\end{table}

\section{Conclusion}
We demonstrate how past knee X-ray exams consisting of sets of images and corresponding radiological reports can be used as part of a learning framework to determine image-text correspondence and hence automate the generation of such reports for new X-ray exams. Preliminary results look promising as the auto-generated reports correlate well with true reports, and we hope to train the model on additional knee X-ray exams as these become available to us. Further developments to the model can be made by incorporating the knowledge of the view-type of each image, keeping them as separate inputs, and finding correspondence between each image and parts of text (for instance, through the use of an attention mechanism \cite{xu2015show}). We will explore this in a future study.

%\subsubsection*{Acknowledgements}

%\section*{References}
\bibliographystyle{unsrtnat}
\bibliography{library} 

\end{document}